\documentclass[conference,letter]{IEEEtran}
\IEEEoverridecommandlockouts
\usepackage{geometry}
\usepackage{cite}
\usepackage{amsmath,amssymb,amsfonts}
\usepackage{algorithmic}
\usepackage{graphicx}
\usepackage{textcomp}
\usepackage{xcolor}
\newcommand{\subsubsubsection}[1]{\paragraph{#1}\mbox{}\\}
\def\BibTeX{{\rm B\kern-.05em{\sc i\kern-.025em b}\kern-.08em
    T\kern-.1667em\lower.7ex\hbox{E}\kern-.125emX}}
\begin{document}

\title{Emergent Multi-View Fidelity in Autonomous UAV Swarm Sport Injury Detection\\

\thanks{This work is partly supported by Research
Ireland, the European Regional Development Fund (Grant No. 13/RC/2077\_P2) and the EU MSCA Project ”COALESCE” (Grant No. 101130739)}
}

\author{\IEEEauthorblockN{Yu Cheng}
\IEEEauthorblockA{\textit{School of Engineering} \\
\textit{Trinity College Dublin}\\
Dublin, Ireland \\
chengyu@tcd.ie}
\and
\IEEEauthorblockN{Harun Šiljak}
\IEEEauthorblockA{\textit{School of Engineering} \\
\textit{Trinity College Dublin}\\
Dublin, Ireland \\
harun.siljak@tcd.ie}
}

\maketitle

\begin{abstract}
Accurate, real-time collision detection is essential for ensuring player safety and effective refereeing in high-contact sports such as rugby, particularly given the 
severe risks associated with traumatic brain injuries (TBI). Traditional collision-
monitoring methods employing fixed cameras or wearable sensors face limitations in visibility, coverage, and responsiveness. Previously, we introduced a framework using unmanned aerial vehicles (UAVs) for monitoring and real time kinematics extraction from videos of collision events. In this paper, we show that the strategies operating on the objective of ensuring at least one UAV captures every incident on the pitch have an emergent property of fulfilling a stronger key condition for successful kinematics extraction. Namely, they ensure that almost all collisions are captured by multiple drones, establishing multi-view fidelity and redundancy, while not requiring any drone-to-drone communication.  

\end{abstract}


\begin{figure*}[t]
\centering
\includegraphics[width=7 in]{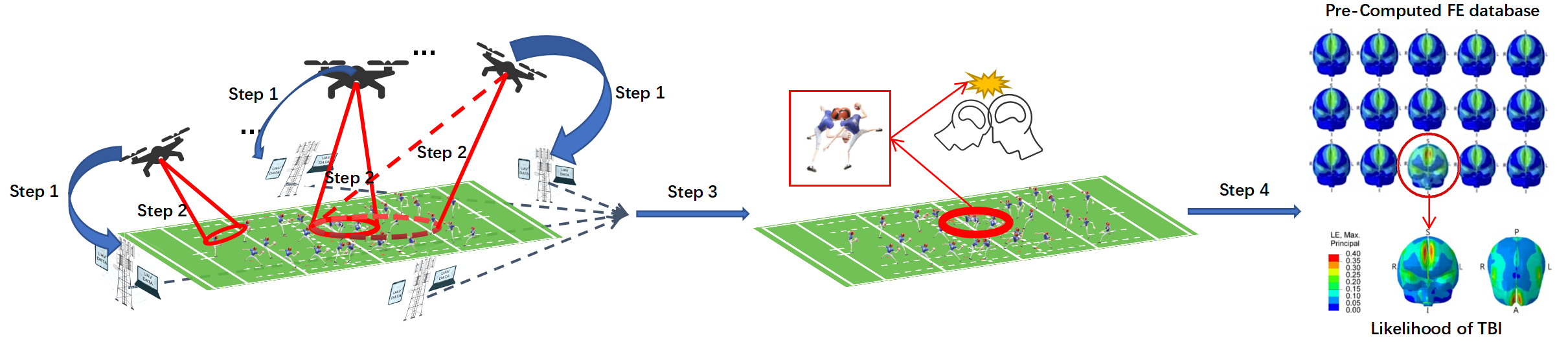}
\caption{The general framework of our system}
\label{system}
\end{figure*}

\section{Introduction}
The growing demand for real-time injury detection in large-scale, dynamic sports environments is driven by the fact that high-impact collisions can cause serious injuries, such as concussions or ligament tears, that often go unnoticed amid the chaotic play in crowded stadiums \cite{tbi0715}. Immediate detection is crucial, as prompt medical response can prevent worsened outcomes \cite{2003realtime}. In fast-paced sports like rugby union, traditional monitoring which is relying on human observation and fixed cameras with post-game reviews, frequently overlooks subtle or off-ball injuries, with studies showing that many concussive events remain undetected by sideline observers \cite{notdetected}. For instance, the King–Devick test identified 77\% (17/20) of concussive incidents unrecognized during live play \cite{notdetected02}, underscoring the need for objective sideline tools to complement human observation.\\
Recent advancements in sports injury monitoring have led to the development of wearable sensor systems, including instrumented mouthguards \cite{Liu_2020} and impact‐detecting patches \cite{Eitzen_Renberg_Færevik_2021}. These support automatically detecting heavy collisions in real time \cite{jones2022ready}, thereby alerting staff to potential concussions by quantifying impact forces \cite{Camarillo_Shull_Mattson_Shultz_Garza_2013}. However, challenges related to coverage, sensor dislodgment, and inherent accuracy issues (e.g., with the X-Patch) often necessitate video review to eliminate false positives and negatives \cite{tooby2024pull}. As an attractive alternative, automated video analysis systems use computer vision to process live feeds from fixed cameras or UAVs to detect injury events, such as a player collapsing without manual intervention \cite{shi-tomas2024}. However, crowded fields and occlusions require multiple points of view, increasing system complexity and data throughput, which can overwhelm available wireless channels and impede real-time responsiveness at stadium scale \cite{videocomm}. Advanced image processing techniques, including those based on the Shi-Tomas algorithm, and multi-camera tracking methods have been developed to enhance feature extraction and improve detection robustness \cite{enhance02}. \\
In live stadium scenarios, limited wireless bandwidth and reliance on frequent inter-drone communication or a central coordinator can introduce latency and complexity, hindering robust redundancy and fault tolerance. To bridge these gaps, one promising direction is deploying autonomous UAV swarms for overhead monitoring, which provide dynamic, multi-angle coverage and real-time capture of collision events. In our previous work \cite{yu}, we developed a decentralized UAV-based collision monitoring system for rugby using an agent-based NetLogo model, demonstrating that a coordinated UAV fleet with onboard collision-detection algorithms outperforms fixed-camera setups by capturing high-impact tackles and potential concussion incidents with greater precision and speed. \\
In this paper, we expand on this concept by proposing an autonomous UAV swarm architecture for injury-event detection that emphasizes robust coordination with minimal communication overhead. Our novel multi-UAV control strategy enables simultaneous, real-time collision detection without explicit drone-to-drone communication; instead, each UAV follows decentralized behavioral rules that naturally foster coverage overlap and redundant observation, thereby preserving bandwidth for essential ground control commands and live video streaming while keeping the UAV network agile and scalable even in communication-constrained environments.
After the introduction of the overall system structure and its objectives in the next section, we define metrics for system performance and define the environment model and self-organising strategies we employ for the swarm. In the results section, we show that methods previously introduced to ensure single-view performance for collisions on the pitch have the emergent property of multi-view fidelity. Finally, in the last section we draw conclusions and point to future work directions.

\section{Methodology}

\subsection{\textbf{System Structure and Objectives}}
The proposed system enables rapid, accurate detection of potential traumatic brain injuries (TBI) in rugby players by integrating autonomous drone surveillance, real-time data processing, and advanced analytical frameworks (see Fig. \ref{system}). In Step 1, drones autonomously capture real-time footage of the match and continuously transmit data to ground control stations for immediate analysis and broadcast. In Step 2, the drones self-organize their flight paths to monitor players for head impacts using automated detection strategies. Step 3 involves fusing multi-drone data at the ground stations, where critical video frames are quickly extracted and analyzed by a reinforcement learning framework trained to identify collision characteristics. Finally, in Step 4, the processed collision data is integrated into a brain analysis model using finite element methodologies \cite{TBI01} to evaluate the likelihood of TBI, with results promptly relayed to facilitate immediate medical response.

\subsubsection{\textbf{Step 1: Real-Time Drone Data Acquisition and UAV Ground Control Station Communication}}
In this initial phase, drones actively transmit real-time positional data, altitude, orientation angles, and high-quality visual streams or images to multiple drone ground control stations strategically placed around the rugby field. Each ground control station integrates functionalities including real-time flight monitoring, dynamic positional plotting on electronic maps, comprehensive mission playback analysis, and continuous antenna tracking for optimized drone tracking and data reception.

\subsubsection{\textbf{Step 2: Autonomous uav swarm Operation and Head Impact Detection}}
Based on predefined autonomous UAV flight strategies, drones coordinate their flight paths independently while continuously monitoring rugby players. The various operational modes employed are detailed in Section C, "Self-organising Strategies". Parameters like drone quantity, speed, and detection radius are adjustable to tailor operations to specific scenarios, enhancing data accuracy. Upon detection of a head impact event, drones immediately initiate data transmission protocols, providing 10 seconds of video footage surrounding the collision event and instantaneous high-resolution images to the corresponding UAV ground control station. Collision detection accuracy is defined mathematically as (\ref{e1}), where $\mathbb{I}_{\mathrm{detect}}(t)$ denotes all true positive impact detections by the UAV swarm and $\mathbb{I}_{\mathrm{total}}(t)$ represents the total validated collisions during the rugby game:
\begin{equation}
\text{Detection Accuracy} = \frac{\sum_{t=1}^T \mathbb{I}_{\mathrm{detect}}(t)}{\sum_{t=1}^T \mathbb{I}_{\mathrm{total}}(t)} \label{e1}
\end{equation}

\subsubsection{\textbf{Step 3: Multi-UAV Data Fusion and Advanced Video Processing for Collision Detection}}
Upon receiving collision event data, drone ground control stations swiftly extract critical frames from the video streams. To accurately identify collisions during rugby matches—specifically, instances of player head impacts—this work employs a minimum of two unmanned aerial vehicles (UAVs) observing each collision event. By requiring that at least two UAVs independently detect a collision at the same location, the subsequent data fusion process is rendered more robust. Once a collision event is detected, ground control stations rapidly extract the relevant frames from the UAV video streams and process them using an advanced reinforcement learning framework integrated with YOLO V8-based object detection. Redundancy is mandated through overlapping spatiotemporal observations—two or more drones must detect the same collision event at the same coordinates and time window. Once a collision is flagged by multiple drones, the associated video frames are forwarded to the UAV Ground Control Station (GCS) for fusion and subsequent analysis.

\subsubsubsection{\textbf{Multi-Drone Collision Reliability Metric}}
To fuse detections from \( N_{\text{drone}} \geq 2 \) drones, a reliability metric is computed using channel-aware weighting. Let \(\text{SNR}_i\) represent the signal-to-noise ratio of drone \(i\)’s video feed, \(\text{DoS}_i\) denote its collision detection confidence score, and \(\gamma_i\) be the YOLOv8 output probability. The overall collision confidence \(\Gamma_{\text{collision}}\) is then:

\begin{equation}
\Gamma_{\text{collision}} 
= \sum_{i=1}^{N_{\text{drone}}} \left( \frac{\text{SNR}_i \cdot \text{DoS}_i}{\sum_{j=1}^{N_{\text{drone}}} \text{SNR}_j \cdot \text{DoS}_j} \right) \gamma_i.
\end{equation}

This approach robustly aggregates observations from multiple vantage points, requiring consensus in both image-based confidence and channel quality before officially declaring a collision event. The weighting mechanism also makes the fusion process adaptive to variations in communication link quality, ensuring that lower-SNR feeds contribute proportionally less to the final decision.

\subsubsubsection{\textbf{Operational Scenarios and Communication Considerations}}
Balancing energy consumption, computation load, and communication overhead is crucial for scalable deployments. Two distinct operational scenarios are considered:
\begin{itemize}
    \item {\textbf{\emph{Edge Processing with Zero Inter-Drone Communication}}}  
    
   Each drone locally runs the YOLOv8 inference for collision detection and calculates the associated metrics (e.g., collision location, force vector, athlete IDs). Since no direct UAV-to-UAV communication occurs, the inter-drone communication load is null. However, onboard computation dominates energy usage, modeled by:
   \begin{equation}
   E_{\text{comp}} 
   = k \cdot f_{\text{CNN}} \cdot t_{\text{proc}} \cdot V_{\text{dd}}^{2},
   \end{equation}

   where \(E_{\text{comp}}\) is the computational energy consumption, \(f_{\text{CNN}}\) denotes the inference frequency of the YOLOv8 model, \(t_{\text{proc}}\) is the total processing time, and \(V_{\text{dd}}\) represents the supply voltage. While this setup offers low-latency detection with minimal reliance on wireless links, it accelerates battery depletion due to intensive onboard inference.

   \item {\textbf{\emph{Cloud Processing with Centralized Computation}}} 
   
   In this paradigm, drones stream raw (or minimally compressed) video data directly to the GCS over orthogonal frequency-division multiple access (OFDMA) channels. Onboard computation is nearly eliminated \(\bigl(E_{\text{comp}} \approx 0\bigr)\), shifting the processing burden to the GCS. However, this approach can potentially saturate network resources as the required bandwidth scales with the number of drones and their video resolutions. The total bandwidth requirement is:
   \begin{equation}
   BW_{\text{total}} 
   = \sum_{i=1}^{N_{\text{drone}}} 
   \Bigl( R_{\text{enc}} \cdot T_{\text{obs}} 
          \cdot \log_2\bigl(1 + \tfrac{P_t \lvert h_i \rvert^2}{N_0 B}\bigr) \Bigr),
   \end{equation}

   where \(R_{\text{enc}}\) is the encoding rate (e.g., H.265), \(T_{\text{obs}}\) the observation window, \(P_t\) the transmit power, \(|h_i|\) the channel gain, and \(N_0 B\) the noise power. Although offloading computations to the GCS conserves drone battery power, surging data rates may exceed 120 Mbps per drone at 4K resolution, risking channel congestion and increased latency.

\end{itemize}

Our proposed framework currently assumes ideal channel conditions, zero packet loss, and ample GCS capacity. Real-world scenarios introduce dynamic wireless channel variations, retransmission overhead, and potential quantization errors. Future research will employ stochastic network calculus to quantify latency-energy tradeoffs under time-varying channels and constrained bandwidth. Additionally, advanced power-saving strategies, such as adaptive inference frequency, partial local pre-processing, will be explored to strike an optimal balance among energy consumption, onboard computation, and communication expenditure. By mandating detection consistency from multiple drones and intelligently partitioning tasks between onboard and centralized processing, this multi-drone collaborative system offers a robust and scalable approach to identifying head collision events in sports analytics.

\subsubsection{\textbf{Step 4: Traumatic Brain Injury (TBI) Likelihood Analysis}}
Critical data from the advanced video analysis stage are subsequently integrated into a sophisticated brain analysis model designed to evaluate the likelihood of traumatic brain injury resulting from observed collisions. Utilizing trained algorithms specifically calibrated for concussion risk assessment based on visual collision parameters, the system promptly calculates TBI likelihood and communicates this information for immediate medical response and intervention decision-making. This step is beyond the scope of this paper.

\subsection{\textbf{The Environment Model}}

Our simulation leverages NETLOGO to unify two key components: a \textbf{Rugby Model} and a \textbf{Drone Model}, into a single, agent-based environment. Each simulation tick, rugby players move according to roles and attributes (speed, passing/shooting preferences), collisions are checked, and drones respond in real time.  

\paragraph{Rugby Model} 
The rugby sub-model replicates standard match elements: a field with realistic dimensions, teams of players (attackers vs.\ defenders, team-oriented vs.\ selfish), ball dynamics (shoot or pass), and scoring rules. Player collisions arise naturally from ball competition or defensive tackles, with these events logged alongside relevant data, such as collision location and severity.

\paragraph{Drone Model} 
Drones act as autonomous agents flying at a fixed height and running various operational modes (\texttt{Fixed}, \texttt{Follow-Ball}, \texttt{Follow-Players}, \texttt{Density-Based}, \texttt{Repulsive}, \texttt{Random}). They either follow the ball, focus on clusters of players, or patrol randomly, adjusting flight paths to avoid collisions and maintain coverage. Whenever a collision is flagged, drones capture and transmit collision frames for potential TBI analysis.

\paragraph{Parameterization and Analysis} 
Building on findings from our previous work, we fix each drone’s detection range at 5 and speed at 10, which are internal model values, relative to speeds and distances in the rugby game, shown to yield best collision monitoring. By standardizing these parameters, we ensure consistently robust detection and facilitate a focused investigation of the multi-UAV data fusion and subsequent TBI likelihood estimation under uniform conditions.

\subsection{\textbf{Self-organising strategies}}

In our previous research \cite{yu}, we proposed six self-organising strategies designed to enable drones to dynamically position themselves for effective collision detection. These strategies operate solely through real-time observation of the game environment, while without relying on drone-to-drone communication or a UAV ground control station. Therefore, it facilitates adaptive positioning based on environmental cues and enhancing collision detection robustness under decentralised conditions.

\begin{itemize}
    \item \texttt{Fixed Mode}: Drones remain at precomputed coordinates determined via past collision hotspots, similar to stationary cameras with less occlusion and calibration issues.
    \item \texttt{Follow-Ball Mode}: Drones orbit the ball at a fixed radius \(R\), adapting positions each tick to maintain continuous coverage of key in-game action.
    \item \texttt{Repulsive Mode}: Extends \texttt{Follow-Ball} by adding collision avoidance. Drones still track the ball but steer away from each other if they draw too close.
    \item \texttt{Follow-Players Mode}: Focuses on “high-risk” players identified through proximity-based metrics. If none qualify, drones revert to \texttt{Follow-Ball}.
    \item \texttt{Density-Based Mode}: Clusters players by location, then assigns drones proportionally to each cluster’s density. Drones form circular patterns around each centroid.
    \item \texttt{Random Mode}: Drones select random positions within the field while maintaining safe distances from neighbors. This provides baseline testing of autonomous movement and collision avoidance.
\end{itemize}

These strategies are devised with the objective of ensuring that every collision is observed by at least one drone. In this paper, we evaluate the hypothesis that, by increasing the number of drones, these strategies will also ensure that each collision is redundantly observed by two or more drones, even without explicit drone-to-drone communication or deliberate coordination mechanisms.

\section{Results}
\subsection{Emergence of multi-view property in tracking strategies}
\begin{figure}[h]
\centering
\includegraphics[width=3.45in]{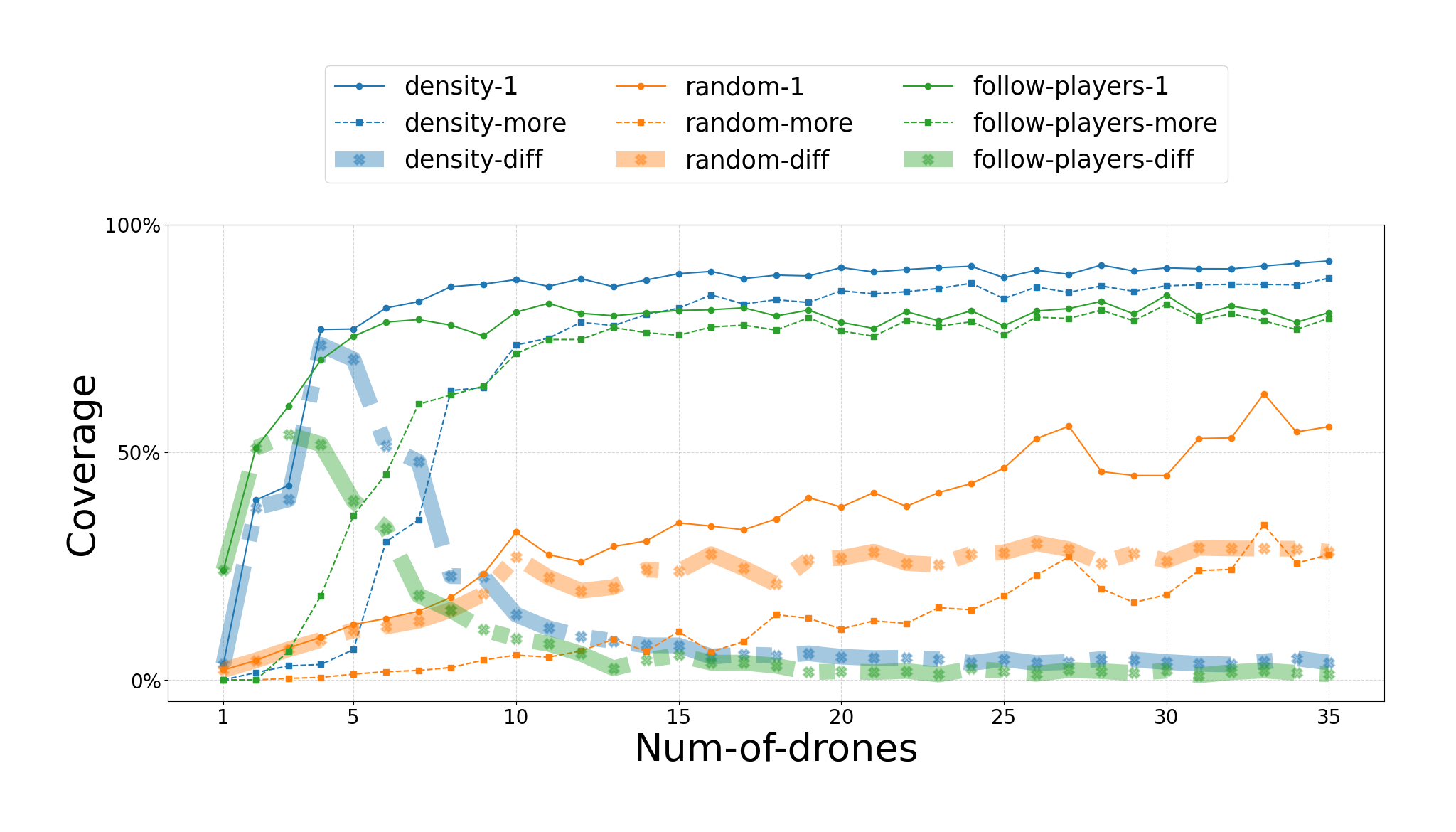}
\caption{Comparative analysis of coverage performance demonstrates averaged results from 100 simulation repetitions conducted in NetLogo. Here, strategy-1 curves indicate number of collisions observed by at least one drone for the given strategy, strategy-more stands for the number of collisions observed by multiple drones (multi-view), and the strategy-diff is the difference between the two, i.e. the number of observed collisions that did not have a multi-view property.}
\label{multiwithdiff}
\end{figure}

In Fig. \ref{multiwithdiff} we compare three collision detection strategies in UAV systems: \texttt{density-based}, \texttt{follow-players}, and \texttt{random} (baseline). Our prior studies\cite{yu} identified density and follow-players as best-performing for single-drone scenarios, while random is retained as a baseline. Results evaluate two metrics per strategy: (1) single-drone detection rate and (2) concurrent detection by $\geq 2$ drones. 

Gap curves quantify performance differences between single and multi-drone modes. The corresponding quantitative performance differences between single-drone and cooperative multi-drone detection modes, including mean differences and maximum positive and negative gap values, are detailed in Table \ref{tab:diff_stats}.

\begin{table*}[htbp]
\caption{Statistical Comparison of Single-Drone vs. Cooperative Multi-Drone Detection}
\begin{center}
\begin{tabular}{|c|c|c|c|c|}
\hline
\textbf{Strategy} & \begin{tabular}[c]{@{}c@{}} Sub-Mode\\ Columns \end{tabular} & \begin{tabular}[c]{@{}c@{}}Mean \\ Difference \end{tabular} &  \begin{tabular}[c]{@{}c@{}} Max\\ Positive\\ Difference \end{tabular} & \begin{tabular}[c]{@{}c@{}} Max\\ Negative\\ Difference \end{tabular} 
\\ \hline

Density &  \begin{tabular}[c]{@{}c@{}} density-1\\ density-more \end{tabular} & 14.76\% & \begin{tabular}[c]{@{}c@{}} +74.70\\ (x=4) \end{tabular}   &  \begin{tabular}[c]{@{}c@{}} -2.45\\ (x=35) \end{tabular}  \\
\hline
Random &  \begin{tabular}[c]{@{}c@{}} random-1\\ random-more \end{tabular}    & 24.72\% &  \begin{tabular}[c]{@{}c@{}} +36.10\\ (x=23) \end{tabular}  &  \begin{tabular}[c]{@{}c@{}} -2.18\\ (x=1) \end{tabular}   \\
\hline
Follow-Players & \begin{tabular}[c]{@{}c@{}} follow-players-1\\ follow-players-more \end{tabular} & 10.44\% &  \begin{tabular}[c]{@{}c@{}} +61.70\\ (x=3) \end{tabular} &  \begin{tabular}[c]{@{}c@{}} -0.70\\ (x=35) \end{tabular}  \\
\hline
\end{tabular}
\label{tab:diff_stats}
\end{center}
\end{table*}

\subsection{Detailed breakdown of multi-view performance}

\begin{figure}[h]
\centering
\includegraphics[width=3.45in]{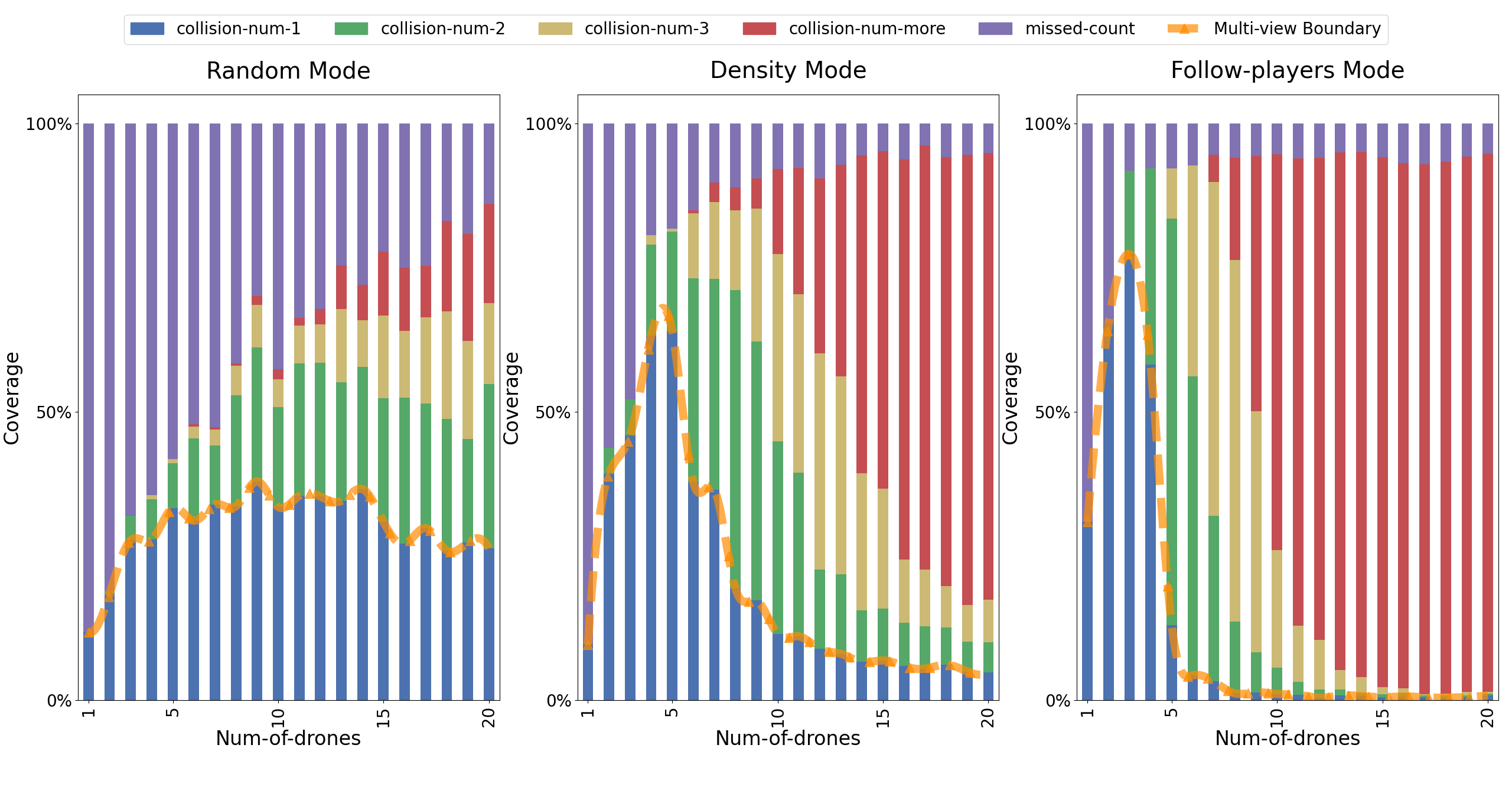}
\caption{Comparative analysis of coverage performance demonstrates averaged results from single instance of strategy execution in NetLogo. Here, collision-num-1, 2, 3, and -more represent percentage of collisions observed by exactly 1, 2, 3 or more UAVs in particular strategy deployments, respectively; missed-count is the percentage of collisions that are not observed at all, and Multi-view Boundary is, akin to diff in Fig. \ref{multiwithdiff}, the number of observations that do not have multi-view property.}
\label{breakdown}
\end{figure}

Although two UAVs theoretically suffice in terms of three-dimensional coverage and kinematics extraction for our purposes, occlusion, poor positioning, and technical issues suggest the need for redundancy. This is why we are interested in the breakdown of drone numbers covering collisions in our experiments. These statistics are shown in Fig. \ref{breakdown} and Table \ref{tab:convergence}.

\begin{table*}[htbp]
\caption{Characteristic Collision Detection Percentages and Convergence Trends}
\begin{center}
\begin{tabular}{|c|c|c|c|c|c|}
\hline
\textbf{Mode} & \textbf{\#Drones} & \textbf{\begin{tabular}[c]{@{}c@{}} collision-num-1\end{tabular} } & \textbf{\begin{tabular}[c]{@{}c@{}} collision-num-2\end{tabular} } & \textbf{\begin{tabular}[c]{@{}c@{}} collision-num-3\end{tabular} } & \textbf{\begin{tabular}[c]{@{}c@{}} collision-num-more\end{tabular} } \\
\hline
Random & 1  & 11.7 & 0.0 & 0.0 & 0.0 \\
\hline
Random & 4  & 27.8 & 7.0 & 0.7 & 0.0 \\
\hline
Random & 10 & 33.5 & 17.3 & 4.8 & 1.8 \\
\hline
Random & 20 & 26.3 & 28.5 & 14.0 & 17.2 \\
\hline
 \begin{tabular}[c]{@{}c@{}} Follow-players\end{tabular}  & 1  & 30.9 & 0.0 & 0.0 & 0.0 \\
\hline
 \begin{tabular}[c]{@{}c@{}} Follow-players\end{tabular}  & 4  & 58.2 & 34.1 & 0.0 & 0.0 \\
\hline
 \begin{tabular}[c]{@{}c@{}} Follow-players\end{tabular}  & 10 & 1.0  & 4.5 & 20.5 & 68.6 \\
\hline
 \begin{tabular}[c]{@{}c@{}} Follow-players\end{tabular}  & 20 & 0.8  & 0.3 & 0.3 & 93.4 \\
\hline
Density & 1  & 9.6  & 0.0 & 0.0 & 0.0 \\
\hline
Density & 4  & 63.2 & 15.7 & 1.7 & 0.0 \\
\hline
Density & 10 & 11.5 & 33.3 & 32.5 & 14.8 \\
\hline
Density & 20 & 4.8  & 5.3  & 7.3  & 77.5 \\
\hline
\end{tabular}
\label{tab:convergence}
\end{center}
\end{table*}


\section{Discussion and conclusion}

From Fig. \ref{multiwithdiff}, the detection accuracy improves as the number of drones increases, though diminishing returns appear beyond a certain threshold. It is also evident that both the detection accuracy and multi-view performance increases much faster for our self-organising strategies, compared to the random baseline. Among all tested strategies, the \texttt{density} mode consistently achieves the highest accuracy. Specifically, \texttt{density-1} surpasses $90\%$ detection accuracy with only five drones and remains stable thereafter. \texttt{Density-more} follows closely, reaching approximately $90\%$ accuracy around seven drones. The \texttt{follow-players} modes exhibit a similar trend, with \texttt{follow-players-1} rapidly increasing in accuracy during the first few drone deployments and then stabilizing. Although \texttt{follow-players-more} remains slightly lower than its single-drone counterpart, its overall performance remains high and stable.

It is interesting to note the difference in multi-view breakdown dynamics of the two self-organising strategies, evident in Fig. \ref{breakdown}. Namely, \texttt{follow-players} exhibits fast convergence to virtually all detected collisions having more than three UAVs observing it (at the 15 drone mark and beyond). The \texttt{density} strategy is slower in that regard, retaining some 1-view, 2-view and 3-view observed collisions throughout. This is still compensated by \texttt{density} performing marginally better at overall collision detection, as evidenced in Fig. \ref{multiwithdiff}.

Therefore, we conclude that strategies designed with the objective of detecting collisions in the rugby game exhibit emergent property of ensuring consistent and reliable multi-view video capture, which improves the quality of kinematics extraction and redundancy in the system. The importance of this emergent feature of the system is twofold: it both allows for simplified design without tracking multiple objectives at design stage, and it guarantees performance without drone to drone communication.\\
Future work in developing this system will focus on the communication performance under realistic conditions.

\bibliographystyle{ieeetr}
\bibliography{IEEE-conference-template-062824/ref}

\begin{thebibliography}{10}

\bibitem{tbi0715}
B.~D. Endres, Z.~Y. Kerr, R.~L. Stearns, W.~M. Adams, Y.~Hosokawa, R.~A. Huggins, K.~L. Kucera, and D.~J. Casa, ``Epidemiology of sudden death in organized youth sports in the united states, 2007--2015,'' {\em Journal of athletic training}, vol.~54, no.~4, pp.~349--355, 2019.

\bibitem{2003realtime}
G.~K. Jonsson, S.~H. Bjarkadottir, B.~Gislason, A.~Borrie, and M.~S. Magnusson, ``Detection of real-time patterns in sports: interactions in football,'' {\em L’{\'e}thologie appliqu{\'e}e aujourd’hui}, vol.~3, pp.~37--45, 2003.

\bibitem{notdetected}
A.~McKinlay and T.~McLellan, ``The mechanism of concussion injury in rugby league,'' {\em INTERNATIONAL SPORTMED JOURNAL}, vol.~15, pp.~328--332, 12 2014.

\bibitem{notdetected02}
D.~King, M.~Brughelli, P.~Hume, and C.~Gissane, ``Concussions in amateur rugby union identified with the use of a rapid visual screening tool,'' {\em Journal of the neurological sciences}, vol.~326, no.~1-2, pp.~59--63, 2013.

\bibitem{Liu_2020}
Y.~Liu, A.~G. Domel, S.~A. Yousefsani, J.~Kondic, G.~Grant, M.~Zeineh, and D.~B. Camarillo, ``Validation and comparison of instrumented mouthguards for measuring head kinematics and assessing brain deformation in football impacts,'' {\em Annals of Biomedical Engineering}, vol.~48, p.~2580–2598, Nov. 2020.

\bibitem{Eitzen_Renberg_Færevik_2021}
I.~Eitzen, J.~Renberg, and H.~Færevik, ``The use of wearable sensor technology to detect shock impacts in sports and occupational settings: A scoping review,'' {\em Sensors}, vol.~21, p.~4962, July 2021.

\bibitem{jones2022ready}
B.~Jones, J.~Tooby, D.~Weaving, K.~Till, C.~Owen, M.~Begonia, K.~A. Stokes, S.~Rowson, G.~Phillips, S.~Hendricks, {\em et~al.}, ``Ready for impact? a validity and feasibility study of instrumented mouthguards (imgs),'' {\em British journal of sports medicine}, vol.~56, no.~20, pp.~1171--1179, 2022.

\bibitem{Camarillo_Shull_Mattson_Shultz_Garza_2013}
D.~B. Camarillo, P.~B. Shull, J.~Mattson, R.~Shultz, and D.~Garza, ``An instrumented mouthguard for measuring linear and angular head impact kinematics in american football,'' {\em Annals of Biomedical Engineering}, vol.~41, p.~1939–1949, Sept. 2013.

\bibitem{tooby2024pull}
J.~Tooby, K.~Till, A.~Gardner, K.~Stokes, G.~Tierney, D.~Weaving, S.~Rowson, M.~Ghajari, C.~Emery, M.~D. Bussey, {\em et~al.}, ``When to pull the trigger: conceptual considerations for approximating head acceleration events using instrumented mouthguards,'' {\em Sports medicine}, vol.~54, no.~6, pp.~1361--1369, 2024.

\bibitem{shi-tomas2024}
Y.~Ma, ``Research on high-definition sports event video image processing system based on computer shi-tomas algorithm,'' in {\em 2024 IEEE 3rd International Conference on Electrical Engineering, Big Data and Algorithms (EEBDA)}, pp.~735--740, IEEE, 2024.

\bibitem{videocomm}
Z.~Shang, J.~P. Ebenezer, A.~C. Bovik, Y.~Wu, H.~Wei, and S.~Sethuraman, ``Assessment of subjective and objective quality of live streaming sports videos,'' in {\em 2021 Picture Coding Symposium (PCS)}, pp.~1--5, 2021.

\bibitem{enhance02}
F.~Yang, S.~Odashima, S.~Masui, I.~Kusajima, S.~Yamao, and S.~Jiang, ``Enhancing multi-camera gymnast tracking through domain knowledge integration,'' {\em IEEE Transactions on Circuits and Systems for Video Technology}, vol.~34, no.~12, pp.~13386--13400, 2024.

\bibitem{yu}
Y.~Cheng and H.~Šiljak, ``Strategies for decentralised uav-based collisions monitoring in rugby,'' Mar. 2025.
\newblock arXiv:2503.22757 [cs].

\bibitem{TBI01}
Q.~Yuan, X.~Li, Z.~Zhou, and S.~Kleiven, ``A novel framework for video-informed reconstructions of sports accidents: A case study correlating brain injury pattern from multimodal neuroimaging with finite element analysis,'' {\em Brain Multiphysics}, vol.~6, p.~100085, 2024.

\end{thebibliography}

\end{document}